\pdfoutput=1
\documentclass{ldr-article}

\usepackage{tabularx}
\usepackage{booktabs}

\newcommand\myTitle{Learning and communication pressures in neural networks: Lessons from emergent communication}

\title{\myTitle{}}

\author{%
Lukas Galke\\
Centre for Machine Learning, Department of Mathematics and Computer Science (IMADA),\\University of Southern Denmark (SDU), Odense, Denmark\\
LEADS group, Max Planck Institute for Psycholinguistics, Nijmegen, Netherlands
\\\\
Limor Raviv\\
LEADS group, Max Planck Institute for Psycholinguistics, Nijmegen, Netherlands\\
Centre for Social, Cognitive and Affective Neuroscience,
University of Glasgow, Glasgow, UK%
}

\usepackage{xspace}
\usepackage{xcolor}
\newcommand\ie{i.\,e.\xspace}
\newcommand\eg{e.\,g.\xspace}

\begin{document}
\maketitle

\begin{abstract}
Finding and facilitating commonalities between the linguistic behaviors of large language models and humans could lead to major breakthroughs in our understanding of the acquisition, processing, and evolution of language. However, most findings on human--LLM similarity can be attributed to training on human data. The field of emergent machine-to-machine communication provides an ideal testbed for discovering which pressures are neural agents naturally exposed to when learning to communicate in isolation, without any human language to start with. Here, we review three cases where mismatches between the emergent linguistic behavior of neural agents and humans were resolved thanks to introducing theoretically-motivated inductive biases. By contrasting humans, large language models, and emergent communication agents, we then identify key pressures at play for language learning and emergence: communicative success, production effort, learnability, and other psycho-/sociolinguistic factors. We discuss their implications and relevance to the field of language evolution and acquisition. By mapping out the necessary inductive biases that make agents' emergent languages more human-like, we not only shed light on the underlying principles of human cognition and communication, but also inform and improve the very use of these models as valuable scientific tools for studying language learning, processing, use, and representation more broadly.
\end{abstract}
    
    \keywords{language acquisition; language evolution; emergent communication;  large language models; learning biases; learning pressures; neural networks; neural language models; multi-agent systems}
    \correspondingauthor{Lukas Galke, Centre for Machine Learning, Department of Mathematics and Computer Science (IMADA), University of Southern Denmark (SDU), Campusvej 55, DK-5230 Odense M, Denmark. Email: galke@imada.sdu.dk}
    \orcid{https://orcid.org/0000-0001-6124-1092}
    \citationinfo{Galke, L. \& Raviv, L. (2024). \myTitle{}. \textit{Language Development Research 5}(1), 116--143. \\\url{http://doi.org/10.34842/3vr5-5r49}}

\section{Introduction}
Using neural language models for language development research dates back to
\textcite{elmanLearningDevelopmentNeural1993} simulating language acquisition
with recurrent neural networks and conceiving the theory of ``the importance of
starting small''. Similarly, \textcite{harrisDistributionalStructure1954}'s
distributional structure has motivated word embeddings -- a seminal work
showing that the semantic relationship between words can be learned without
supervision from text data
alone~\parencite{mikolovDistributedRepresentationsWords2013,gothDeepShallowNLP2016}.
These are just some examples of where machine learning has already influenced
the development and testing of linguistic theories, showcasing a thriving
relationship between the two
disciplines~\parencite{deseysselRealisticBroadscopeLearning2023,dupouxCognitiveScienceEra2018,baroni2021proper,contreraskallensLargeLanguageModels2023a}.
The unprecedented success of language models in recent
years~\parencite{bahdanauNeuralMachineTranslation2015,vaswani_attention_2017,bert,t5,gpt3}
provides many opportunities to further advance our understanding of human
language learning. 

A growing body of work has found similarities between large language models and
humans~\parencite{dasguptaLanguageModelsShow2022,webbEmergentAnalogicalReasoning2023,schrimpfNeuralArchitectureLanguage2021,srikant2022convergent,wei2022emergentAbilities},
showing that approximate representations of the outside world can be learned
from statistical patterns found in linguistic input
alone~\parencite{li2022emergent,abdouCanLanguageModels2021,liImplicitRepresentationsMeaning2021,patelMappingLanguageModels2022},
and manifesting the usefulness of large language models for other disciplines
such as psychology~\parencite{demszkyUsingLargeLanguage2023}. However, a so far
open issue is the fact that language models are exposed to different input
modalities (i.e., mainly text) and have much more data available for training
than
humans~\parencite{deseysselRealisticBroadscopeLearning2023,warstadtWhatArtificialNeural2022}.
Resolving the discrepancy by which language models require much more data than
a human child is of high interest to both cognitive science (with the goal of
more representative models) and natural language processing researchers (with
the goal of more efficient models). Notably, there are ongoing efforts to train
language models from similar input as available to a human child, \eg, as in
BabyBERTa~\parencite{huebner-etal-2021-babyberta}, and the BabyLM
challenge\footnote{\url{https://babylm.github.io}}~\parencite{warstadt2023papers}.

To promote a deeper understanding of how large language models may
be useful for language development research, we suggest to take
inspiration from the field of emergent machine-to-machine communication -- where two or more neural network agents without exposure to an existing language need to engage in a communication game with the
goal of successfully understanding each
other~\parencite{foersterLearningCommunicateDeep2016,kottur-etal-2017-natural,DBLP:conf/iclr/LazaridouPB17,lazaridou2020emergent}. Specifically, emergent communication simulations explore what happens when artificial neural networks (on which also large language models are based) need to create their own languages from scratch, i.e., without first being pre-trained on natural language corpora: do they create human-like languages by-default, or are there specific biases and constraints that need to be introduced in order to replicate human behavior? By attempting to simulate phenomena previously observed in humans, research on emergent communication has provided valuable insights into the processes and pressures that shape the evolution of human language, and has allowed researchers to effectively scrutinize, identify, and tease apart the relevant learning biases and conditions that underlie the communicative behaviors of artificial neural networks when they are made to communicate by themselves.

Although the setting of emergent communication is typically motivated for studying the evolution of language~\parencite[see][inter alia]{lazaridou2020emergent,lianCommunicationDrivesEmergence2023a},
language learning and language evolution are intrinsically linked: As languages
are passed from generation to generation in a repeated cycle of transmission,
imitation, and use, their structure is continuously shaped by the pressures and
biases introduced by learners during the process of language acquisition --
with such learning biases effectively shaping the evolution of languages on a
longer timescale~\parencite{chaterLanguageAcquisitionMeets2010,kirby2014iterated,smith2022language}.
As such, constraints and pressures associated with learning can causally affect
(and, in fact, create) the universal properties of languages, including their
most fundamental structural
features~\parencite{kirby2002learning,kirby2004ug,kirby2017culture}. As such, we believe that the field of emergent communication provides an ideal testbed for exploring the learning pressures neural networks are exposed to in the process of language learning and use, and can help shed light on (some of) the criticial inductive biases needed for replicating human linguistic behavior. 

Since the theoretical usefulness of a model is dependent on its resemblance to the target entity ~\parencite{zeigler2000theory}, identifying the relevant learning pressures and biases that govern language creation in neural network models can in turn make neural language models more behaviorally plausible, and consequentially a more robust scientific tool for the language sciences. 
Here, we review the emergent communication literature and  identify underlying learning pressures, while contrasting those with the learning pressures at play when training large language models. Thereby we shed new light on the learning dynamics of neural language models and contribute to the development of more behaviorally plausible language models for language acquisition research. 

In the following, we offer a comparative perspective on humans, large language models, and deep learning agents engaging in communication games by reviewing similarities and differences in observed phenomena, discussing how mismatches in the behavior of humans and neural agents can be resolved through appropriate inductive biases, and determining the underlying learning pressures at play. We first provide a brief overview of the emergent
communication literature, and then showcase initial mismatches between neural
agents and humans with respect to multiple linguistic phenomena: Zipf's
law of abbreviation, the benefits of compositional structure, and social
factors shaping linguistic diversity (e.g., population size effects). For each of these phenomena, we describe how the initial mismatch between humans and neural network models has been resolved, and identify the underlying learning pressures giving
rise to these patterns. In particular, we identify four cognitive and
communicative pressures underlying both language acquisition and language
evolution, and discuss whether they are inherent to the training objective (i.e., present by default given the learning environment and objective) or
whether they need to be artificially incorporated into the models as inductive
biases to elicit the desired outcome. We then contrast the identified pressures and biases with those present in the training of large language models, with the goal of promoting knowledge transfer between machine learning and language sciences. We conclude with concrete suggestions for future directions, aimed at developing more cognitively plausible language models for both language development and language evolution research.

\section{Emergent communication, initial mismatches, and their resolution}
\begin{figure}[ht!]
    \centering
    \includegraphics[width=0.6\textwidth]{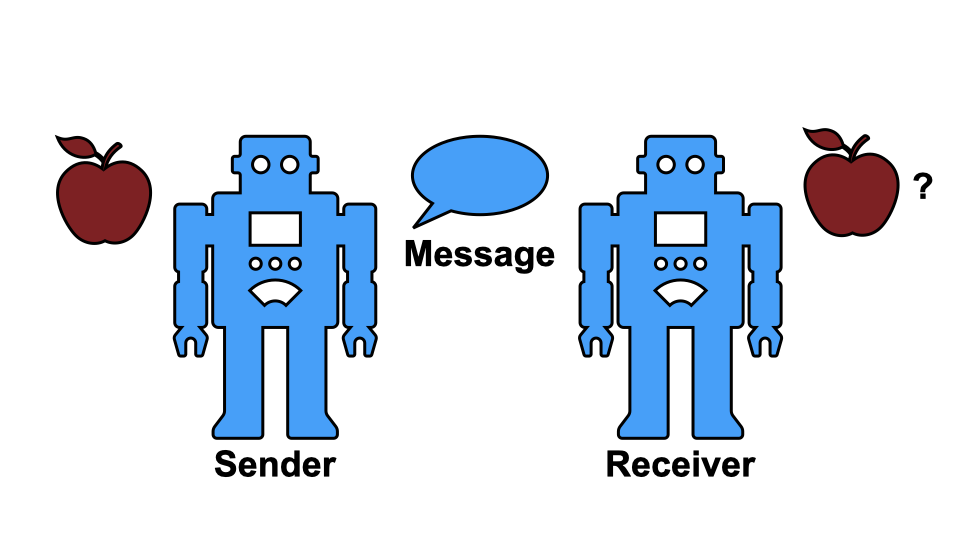}
    \caption{Schematics of a simple communication game. The sender sees an object and has to compose a message to describe it. The receiver only sees the message and has to discriminate the object against distractors, or fully reconstruct it.}\label{fig:commgame}
\end{figure}

\begin{table}
    \caption{Observed phenomena from humans in agents from emergent communication simulations}\label{tab:phenomena}
    \begin{tabularx}{\textwidth}{|p{4cm}|X|X|}
        \hline
        \textbf{Phenomenon in Humans} & \textbf{Mismatch in Emergent Communication agents} & \textbf{Resolution}\\
        \hline
        Zipfian distribution in utterance length (frequent meanings are described by shorter utterances) & Sender agents exploit the full channel capacity because longer messages are easier to distinguish by receiver agents. 
        & Introducing a penalty on long utterances (simulating "laziness") restores the Zipfian distribution on utterance length.\\
        \hline
        Compositional structure reliably emerges during communication and cultural transmission, and is beneficial for language learning and generalization & Inconsistent emergence of compositional structure in neural agents, and seemingly no advantage of more compositional protocols for generalization &
        Periodically resetting agents' parameters (simulating generational turnover) gives rise to compositional protocols, which are easier to learn for neural network agents\\
        \hline
        Population size affects the emergence of compositional structure (larger communities create more systematic languages) & Larger populations of neural agents do not create more compositional protocols & Introducing population heterogeneity (simulating individual differences) or production-comprehension symmetry (simulating role alternation in language use) leads to larger populations creating more systematic protocols\\
        \hline
    \end{tabularx}
\end{table}

Computational modeling has long been used to study
language evolution by simulating the process of communication and transmission between artificial agents, typically Bayesian learners
\parencite{kirby2002learning,smith2003complex,kirby2004ug,smith2008cultural,gong2008exploring,dale2012understanding,perfors2014language,kirby2015compression,steels2016agent}. The emergence of new communication systems is similarly studied using deep
neural network models~\parencite{lazaridou2020emergent}, and in experimental
work with human
participants~\parencite{selten2007emergence,kirby2008cumulative,winters2015languages,raviv2019larger}.
Regardless of whether the subjects of these experiments are humans, Bayesian
agents, or deep neural networks, they all share the same methodological
framework, namely, sender-receiver communication games: One agent describes an
input (\eg, an object or a scene), and transmits a message to another agent,
that then has to guess or fully reconstruct the sender's input (see
Figure~1). The agents in emergent communication experiments are
typically based on deep neural networks, similar to those used in large language models.

In a typical communication game, the sender acts as a conditioned-generation
model, taking a target input (for example, an image or a set of attribute
values) and produces a message consisting of multiple symbols. The symbols of
the message are generated one by one without any pre-defined vocabulary. The
generated message is then transmitted to the receiver. The receiver is trained
to infer the sender's input based on the message, by selecting
the correct object among distractors or by fully reconstructing it. 

Emergent communication models start with randomly-initialized
parameters, without any pre-defined list of words or look-up table. Thus, the
messages start out as random, and only over the course of training and
interaction do the models develop a communication protocol. In fact, it is the
central assumptions of \emph{emergent} communication that the agents are not
seeded with some initial language or communication protocol, but that they
develop the communication system on their own during interaction. Thus, agents
start from scratch and are guided primarily by communicative success. Yet,
there is room for inductive biases, \ie, additional biases that are imposed on
the learning system to promote desired
behaviours~\parencite{mitchellNeedBiasesLearning}. While cognitive biases in
biological learning systems occur naturally, inductive biases in machine
learning are artificially introduced to guide the learning dynamics. For a
profound overview of the emergent communication literature, we refer to recent
review and survey papers by \textcite{lazaridou2020emergent},
\textcite{galke2022emergent}, and \textcite{brandizziMoreHumanlikeAI2023}.

Notably, methods from the field of emergent communication and from the closely related field of reinforcement learning~\parencite[see][inter alia]{kosoy2020exploring,pmlr-v177-kosoy22a} have already been used for language development research~\parencite[see][inter alia]{ohmer2020reinforcement,portelance2021emergence}, for example, to study the emergence of a mutual exclusivity bias with pragmatic agents.

While emergent communication simulations hold a great potential for advancing
our understanding of how languages emerge, we can only expect insights gained
with deep neural networks to inform language evolution research if the
resulting languages actually show the same properties as natural
languages~\parencite{galke2022emergent}. Consequently, most emergent
communication simulations try to compare the properties of their emerging
communication protocols to the properties found in natural
languages~\parencite{DBLP:conf/iclr/LazaridouPB17,DBLP:conf/nips/HavrylovT17,kottur-etal-2017-natural}.
By following this approach, the field has unveiled substantial differences
between humans and machines in how they learn to communicate and what kinds of
languages they develop.

Crucially, although the emergent languages of neural networks initially did not
exhibit many of the linguistic properties typically associated with human
languages, most of these differences could be reconciled by adding adequate
inductive biases, such as laziness and impatience -- which, when introduced,
recovered the effects found in humans. Notably, some linguistic phenomena such as the 
word-order/case-marking trade-off seem to occur in communicating neural
networks without specific inductive
biases~\parencite{lianCommunicationDrivesEmergence2023a}.
Below we review selected properties of
human languages in which initial mismatches between humans and neural network
agents were resolved and discuss the inductive biases that were necessary for
their recovery. Table~\ref{tab:phenomena} provides an overview of the three phenomena and their occurrence in neural simulations.

\subsection{Zipfian distribution in utterance length}

Perhaps the most
illustrative example of mismatches between the languages developed by humans
and machines was the initial absence of Zipf's law of abbreviation in machine
learning simulations. According to Zipf's law of abbreviation, the relationship
between word frequency and word length follows a power law distribution, such
that more frequent words are typically shorter while less frequent words are
typically longer~\parencite{ZLA,newman2005power}. \textcite{ZLA} suggested that
this effect is caused by the principle of least effort, i.e., since frequent
words are produced often, and shorter words are easier to produce. Critically,
Zipf's law has important implications for language
evolution~\parencite{kanwal2017zipf} and language
acquisition~\parencite{ellisInputSecondLanguage2009}, with active restructuring
of lexicon towards more efficient
communication~\parencite{gibsonHowEfficiencyShapes2019a}.

Initial findings in emergent communication showed that
Zipf's Law of Abbreviation is absent from the languages developed by neural
agents, which was dubbed as 'anti-efficient
coding'~\parencite{chaabouni2019anti}. This was because neural senders were not
under any pressure to communicate efficiently or to reduce effort. In fact,
longer messages were easier for the receiver agent to process because they
allowed for more opportunities to differentiate between meanings: for a
1-symbol utterance, the sender can select only $1$ item from the alphabet of
size $k$, but for a $n$-symbol utterance, the sender can produce $k^n$
different combinations. The more distinct utterances are from another, the
easier it is for the receiver to distinguish the target meaning from other
possible meanings. Thus, longer utterances are advantageous for conveying the
meaning correctly -- especially when there is no penalty for utterance length. 

The mismatch with human language was resolved by
adjusting the optimization objective in a direction that made sender agents
``lazy'' (i.e., longer messages were penalized) and receiver agents
``impatient'' (i.e., receivers tried to infer the meaning as early as possible
in a sequential read)~\parencite{DBLP:conf/conll/RitaCD20}. This inductive
bias, which aims at mimicking real human behavior during language production
and comprehension, has recovered Zipf's Law of Abbreviation in emergent
communication simulations -- showing that when such biases for efficiency are
introduced, communication protocols developed by neural agents do show a
similar frequency--length relationship as found in natural languages.

\subsection{The emergence of compositional structure and its benefits for learning and generalization}

Compositional structure is
considered a hallmark feature of human
language~\parencite{hockett1960origin,szaboCompositionality2022}: there is a
systematic mapping between linguistic forms (e.g., words, morphemes) and their
meanings (e.g., concepts, grammatical categories), such that the meaning of a
complex expression can be typically derived from the meanings its constituent
parts. For example, the meaning of the phrase "small cats" is directly derived
from the meanings of the words "small", "cat", and the marker "-s" (denoting
plurality). The presence of such compositional structure underlies the infinite
expressive and productive power of human languages, allowing us to describe new
meanings in a way that is transparent and understandable to other
speakers~\parencite{kirby2002learning,zuidema2002poverty}.
 
In experiments simulating the evolution of languages in the lab using
sender-receiver communication games, the need to communicate over a growing
number of different items or in an open-ended meaning space leads to the
emergence of compositional languages
~\parencite{nolleEmergenceSystematicityHow2018,raviv2019compositional}.
Crucially, the degree of compositional structure in linguistic input then
predicts adults' learning and generalization accuracy, such that, compared to
languages with little to no compositionality, languages with more compositional
structure are learned better and faster and result in better (i.e., more
transparent and systematic) generalizations to new meanings, which are also
shared across different individuals who never interacted before
~\parencite{raviv2021makes}. Thus, the evolution of more compositional and
systematic linguistic structure allows for more productive generalization and
facilitates communication and convergence between strangers.
 
The learning advantage of more compositional structure for adult participants is also echoed in numerous iterated learning studies, which have shown that
artificial languages become more compositional and consequently easier to learn
over the course of cross-generational
transmission~\parencite{kirby2014iterated,carr2017cultural,kirby2008cumulative,beckner2017emergence}.

Testing the limits of our imagination, neural networks seemed to generalize well
even without compositional communication
protocols~\parencite{DBLP:conf/acl/ChaabouniKBDB20,DBLP:conf/iclr/LazaridouHTC18}.
Specifically, \textcite{DBLP:conf/acl/ChaabouniKBDB20} found that, after
many repetitions of an emergent communication experiment, all compositional
languages generalized well, but so did non-compositional languages. This finding
spurred numerous follow-up studies that aimed at improving the learning dynamics
through inductive biases or by making the communication game more difficult
(more complex stimuli, larger alphabet, longer messages, more agents) to
successfully promote the emergence of compositional
structure~\parencite{ritaEmergentCommunicationGeneralization2022,chaabouni2022emergent}.
However, the lack of correlation between the degree of compositional structure
-- as measured by topographic similarity~\parencite{DBLP:journals/alife/BrightonK06}
-- and generalization performance had remained.

The most reliable way to promote the emergence of compositional languages is
periodically resetting the parameters of the neural network
agents~\parencite{DBLP:conf/nips/LiB19,zhouFortuitousForgettingConnectionist2022,chaabouni2022emergent},
similar to \textcite{kirby2014iterated}'s iterated learning paradigm -- leading
to the hypothesis that compositional languages have a learnability
advantage~\parencite{DBLP:conf/nips/LiB19,guo2019emergence,DBLP:conf/acl/ChaabouniKBDB20,chaabouni2022emergent}.
However, these attempts did not directly test language learnability in a purely
supervised fashion.

Recently, \textcite{conklinCompositionalityVariationReliably2022c} have re-analyzed the setting of \textcite{DBLP:conf/acl/ChaabouniKBDB20} and found that, in fact,
the lack of correlation between compositionality and generalization performance
in the original simulation was caused by a fallacy of the topographic similarity metric that had been
used to measure compositionality. For instance, homonyms
(different forms for same meaning) obscure compositionality under the
topographic similarity measure. When taking this variation into account,
compositional structure does reliably emerge and is beneficial for
generalization. In other words, it is probably the case that there
was not really a mismatch between humans
and neural agents in the first place.

Supporting this view, \textcite{galke2023e2dl} have replicated a large-scale
language learning study originally conducted with human
participants~\parencite{raviv2021makes} with deep neural networks and have
confirmed the advantage of compositional structure for learning and
generalization in neural networks. The results showed similar pattern across
three learning systems -- humans, small-scale recurrent neural networks trained
from scratch, and the large pre-trained language model GPT-3 -- with
compositional structure being advantageous for all types of learners.
Specifically, the results showed that neural networks benefit from more
structured linguistic input, and that their productions become increasingly
more similar to human productions when trained on more structured languages.
This structure bias can be found in the networks' learning trajectories and
their generalization behavior, mimicking previous findings with humans:
although all languages can eventually be learned, languages with a higher
degree of compositional structure were led to better and
more human-like generalization to new, unseen items.

\subsection{Population size effects}
Socio-demographic factors such as population size have long been assumed to be important
determinants of language evolution and variation
\parencite{wray2007consequences,nettle2012social,lupyan2010language}.
Supporting this idea, global cross-linguistic studies report that bigger
communities tend to have languages with more regular and transparent structures
\parencite{lupyan2010language}. Similarly, in experimental work, larger groups
of interacting participants generally develop languages with more systematic
(i.e., compositional) grammars ~\parencite{raviv2019larger}. These findings are
typically attributed to compressibility pressures arising during communication:
remembering partner-specific variants becomes increasingly more challenging as
group size increases and shared history decreases, which lead larger groups to
prefer easier-to-learn-and-generalize variants and thus converge on more
transparent and systematic languages.

\textcite{tieleman2019shaping} has investigated populations of autoencoders.
Autoencoders are neural network models composed of an encoder module and a decoder
module that learn to ``good'' representations (the code) by reconstructing their own
input. Now \textcite{tieleman2019shaping} have 
decoupled  encoder and decoders and exchanged them throughout training
-- while communicating in a continuous channel. There, larger communities produced
representations with less idiosyncrasies and lead to better convergence among different agents.
While a promising starting point, the communication was modeled as exchanging
continuous vectors and training the encoder decoder modules together, as if
they were one model.
This is arguably natural communication paradigm for neural networks
because it is optimized in the same way as the communication between layers in a single neural network.
However, this continuous channel stands in contrast with the discrete nature of human
communication~\parencite{hockett1960origin}. Most other approaches in emergent communication, however, do consider a discrete channel~\parencite{galke2022emergent}.

While \textcite{chaabouni2022emergent} argued that it is necessary to scale up
emergent communication experiments in different aspects including population
size in order to better align neural emergent communication with human language
evolution, they have not found a consistent advantage of population size in
generalization and ease-of-learning (in contrast with
\parencite{tieleman2019shaping}). Similarly, \textcite{rita2022on} found that
language properties are not enhanced by population size alone. 

While emergent communication in
populations of agents has been investigated earlier~\parencite[e.g.]{fitzgerald2019populate,DBLP:conf/emnlp/GraesserCK19,lowe2019learning}, the effect of
population size on structure with groups of more than two agents has only
recently been
analyzed~\parencite{chaabouni2022emergent,rita2022on,michelRevisitingPopulationsMultiagent2023}.
Out of these, two studies aimed to recover the group size effect in populations
of neural network agents by introducing population
heterogeneity~\parencite{rita2022on} and manipulating sender-receiver
ties~\parencite{michelRevisitingPopulationsMultiagent2023}.The first study by \textcite{rita2022on} modeled population heterogeneity by giving each agent a different random learning rate
While previous simulations used populations of identical agents, Rita et al. modeled population heterogeneity by giving each agent a
different random learning rate.  Results showed that in this scenario, group size
effects could be partially recovered. Notably, the authors found that it is
important to give sender agents having (much) higher learning rates than
receivers. 

Secondly, while most emergent communication simulations keep senders and receivers
distinct (i.e., agents that produce never comprehend, and vice versa),
there is also work that emphasizes linking production and comprehension components within the agents
(e.g., by sharing some of the model parameters)~\parencite{DBLP:conf/emnlp/GraesserCK19,portelance2021emergence}.
\textcite{galke2022emergent} argue that this naturalistic property of alternating between sending and
receiving (i.e., engaging in both production and comprehension in typical language use) may be a crucial ingredient to ensure more linguistically plausible
learning dynamics -- and could lead to recovering the group size effect. Subsequently, \textcite{michelRevisitingPopulationsMultiagent2023} have introduced sender-receiver ties via gradient blocking,
such that a sender and a receiver together form a single agent and each receiver
is only optimized for its corresponding sender. This change indeed led to a recovery of the
group size effect, with larger population of agents creating more compositional
protocols. Another promising approach is to have agents model other agents' knowledge, allowing them to communicate differently with different agents - something that has been implied to underlie group size effects in humans~\parencite{meir2012influence,thompson2020complexity,mudd2020agent,lutzenberger2021formal}. While such "theory of mind" is generally absent from emergent communication simulations in populations, the ability to infer other agents' beliefs has been successfully implemented in various reinforcement learning setups, e.g., \parencite{filos2021psiphi,ohmer2020reinforcement}.

\section{Underlying learning pressures and inductive biases}\label{sec:pressures}

\begin{table}
    \caption{Pressures derived from emergent communication simulations and their operationalization in neural agents and large language models}\label{tab:pressures}
    \begin{tabularx}{\textwidth}{|p{4.0cm}|X|X|}
        \hline
        \textbf{Derived Pressure} & \textbf{Emergent Communication Agents} & \textbf{Large Language Models}\\
        \hline
        Pressure for successful communication & The main training objective in communication games & Absent in pre-training and fine-tuning. Only introduced when learning from human preferences in RLHF.\\
        \hline
        Pressure for learnability & Can be artificially introduced through parameter reset and iterated learning  & Neural networks underlying large language models have a tendency to find the simplest solution first \\
        \hline
        Pressure to reduce production effort & Can be artificially introducing, e.g.,  through a penalty term for long messages & Production length is learned from LLM's training data and human feedback in RLHF.\\
        \hline
        Memory constraints & Absent because the high capacity of neural agents is sufficient to memorize even unstructured mappings & Huge capacity due to extremely high amount of parameters, yet ``working memory'' for in-context learning is limited by context window (how many tokens the models can process at a time)\\ 
        \hline
        Production-comprehension symmetry & Can be artificially introduced by linking sender and receiver modules & By design -- LLMs employ the same neural network modules and parameters for comprehension and production\\ 
        \hline
        Modeling other agents' internal states & Can be modeled explicitly, e.g., for pragmatic reasoning & In the RLHF training stage, a reward model is trained and consulted to estimate human preferences. \\
        \hline
    \end{tabularx}
\end{table}

In general, there are two types of learning biases and pressures.
First, some biases and pressures seem to be present naturally, or universally,
across all different learning systems investigated here, including deep learning agents. An example
for this is the structure-bias, i.e., the learnability and generalization
advantage of more compositional communication protocols~\parencite{galke2023e2dl}
(see above). This structure-advantage seems to be present for both humans and neural networks, even without specific inductive biases. 
In contrast, some biases need to be artificially introduced in order
to recover the effects found in humans. These include, for example, adding a
length-penalty for senders, which effectively makes agents "lazy".
In the above examples, we demonstrated the flexibility and adaptive nature of
neural simulations and how they can be tweaked to replicate human behavioral
patterns. While many features associated with natural languages were initially
absent from such simulations, these mismatches have been fully or partially
resolved by introducing theory-driven and human-inspired cognitive biases and
learning pressures to the learning system -- and these inductive biases have
consequentially led to better alignment between neural agents and humans.
Below, we outline on a more fine-grained level what pressures are relevant for
language learning and evolution in neural networks, contrasting them with the
pressures to which current large language models are exposed, and to what
extent incorporating the pressures may promote the relevance of large language models for developmental
research.
Table~\ref{tab:pressures} provides an overview of the comparison of learning pressures in emergent communication agents and large language models.
Notably, this is not an exhaustive list -- it focuses on the specific
pressures that underlie the phenomena described above, but do not consider many
other important aspects that govern natural language learning, such as grounding, a noisy
environment, multi-modal communication, or referential and iconic signs.

\subsection{Pressure for successful communication}
In order to achieve successful communication, language users need distinguish between a variety of
meanings. This expressivity pressure is hypothesized to underlie human language
evolution, and serves as a "counter pressure" for simplicity/compressibility
(i.e., the idea that languages should be as simple and as learnable as
possible)~\parencite{kirby2015compression}. The pressure for communicative
success, \eg, to accurately reconstruct the meaning of referents from a message
during interaction, is the most straight-forward pressure found in
collaborative communication games (and, arguably, in real-world interaction).
In emergent communication with deep neural networks, this pressure is encoded
right in the optimization objective of the neural networks.

In contrast, for large language models such as GPT-3.5, the main objective during pre-training is not communication success. The standard
language modeling objective used during pre-training of large language models instead optimizes for
utterance completion (\ie, learning to predict words from their context).
While this language modeling objective leads to tremendous success regarding
language competence other emergent
abilities~\parencite{bert,wei2022emergentAbilities}, it is clearly a different
training objective than optimizing for communicative success, as in
emergent communication simulations. After large-scale pre-training, large language models are fine-tuned using small datasets of human-generated pairs of
instructions and their corresponding responses, usually with the same training objective as in pre-training. In other words,
the models are made to learn from interactions by completing utterances from human-generated interactions, but not by interacting themselves.
Only during the last stage of training, the models are trained via Reinforcement Learning from Human Feedback (RLHF), where a reward model  estimates human preferences based on human ratings of
different machine-generated responses~\parencite{schulman2017proximal,rlhf}. Only in this final
RLHF training stage of LLMs, the models are optimized for successful
communication. Yet, this stage is important to turn base models
into chat assistants that engage in conversations with humans~\parencite{rlhf,gpt4}.

In general, while emergent communication simulations are tuned for communicative success by design, this
is in fact an extra step in large language models after pre-training on utterance completion.
Thus, the learning paradigms of fine-tuning and subsequent learning from human feedback are worth further exploration for the goal of having
language models being more representative of human behavior. For instance, a
recent study has showcased that fine-tuning large language models on data from psychological
tests turns them into useful cognitive models~\parencite{binz2023turning}.

\subsection{Pressure to reduce production effort}
Humans constantly strive to reduce effort during
interaction~\parencite{gibsonHowEfficiencyShapes2019a}. For instance, this is
demonstrated by our tendency to shorten or erode highly frequent words
~\parencite{ZLA,kanwal2017zipf}. However, the pressure to communicate with least
effort is absent in neural networks, and is usually not reflected in their
training objective. In other words, it simply does not cost more ``effort'' for
a neural network to generate a longer message. By introducing a bias for more
efficient communication, \textcite{DBLP:conf/conll/RitaCD20} have shown that
typical human behavior can be recovered. Since language models similarly don't
have an 'innate' pressure to reduce effort, it may be worth considering
integrating such a pressure for efficient communication into these models for
the sake of mimicking human behavior with respect to language development. 
However, one needs to strike a balance, as imposing a least-effort bias
could also lead to communication failure in emergent communication
scenarios~\parencite{lianEffectEfficientMessaging2021}, calling for further
investigation of how a least-effort bias is best incorporated.

In large language models, there is no pressure to reduce
production effort: LLMs are trained on next-token production over large corpora of text data,
which is being piped through the model in a batched fashion to maximize throughput~\parencite[see for instance][inter alia]{gpt3,touvron2023llama}.
Thus, the main driver for production length is simply the utterance length in data, and the placement of specific separator tokens, e.g., at the end of each unit of consecutive text during training.
 Moreover, the RLHF stage of training large language models~\parencite{rlhf,schulman2017proximal}, which is supposed to align LLMs with human preferences, even promotes the generation of longer utterances, as they are deemed to be more ``helpful'' by (instructed) human annotators~\parencite{singhal2024longwaygoinvestigating}.

At inference time, when the LLM is prompted to generate text, a hard
cut-off on the number of tokens or a soft length penalty may be introduced --
the details of these techniques, however, are often not publicly available.
 Regardless, the training procedure itself does usually not include a length penalty, which needs to be taken into account when planning to use large language models for language development research.

\subsection{Pressure for learnability}

Based on our review, a pressure for learnability (or continual re-learning)
also governs the development of communication protocols between neural network
agents. That is, agents should prioritize communication protocols (or single
variants) that are easier to learn, and such protocols should in turn boost
performance.  This learnability pressure is strongly connected to the fact that
languages must be transmitted, learned, and used by multiple individuals, often
from limited input and with limited exposure time~\parencite{smith2003complex}.
Yet, there is a subtle difference to strict transmission chains of iterated
learning, as it is sufficient with neural networks to reset only some of the
agents~\parencite{DBLP:conf/nips/LiB19}, or only parts of a single
agent~\parencite{zhouFortuitousForgettingConnectionist2022}.  In numerous
different settings, it has been shown that learnability pressures are crucial
for compositional structure to
emerge~\parencite{DBLP:conf/nips/LiB19,chaabouni2022emergent,zhouFortuitousForgettingConnectionist2022}.

This also suggests that under repeated learning, either in Iterated Learning
with human participants or with parameter reset in neural networks, weak
learning biases can get amplified in the process of cross-generational
transmission~\parencite{realiEvolutionFrequencyDistributions2009a}.  But what
are these learning biases exactly? How can they be operationalized? And how do
they actually translate into language learning in the real-world? For example,
do these biases differ between children and adults, or between different levels
of linguistic analyses (e.g., vocabulary vs. syntax)? At the moment, these are
still open questions. However, they highlight the need to seriously consider
the meaning and implications of different modeling choices when simulating
language acquisition using language models and deep neural networks. 

As for large language models, \textcite{chen2024sudden} have made relevant
findings by analyzing the learning dynamics: language models pick up grammar as
the simplest explanation for the data very early on during training (structure
onset), and only shortly thereafter, general linguistic capabilties arise.  In
addition, when suppressing grammar as a possible way to explain the data, the
models learn other strategies, but do not go back to grammar when the
constraint is removed later in training.

This finding connects well with more general findings of simplicity bias in neural
networks~\parencite{geirhos2020shortcut}.
In addition, it also connects with the
findings of emergent communication in emphasizing that re-learning (\eg, through
parameter reset) is important for compositional structure to emerge~\parencite{DBLP:conf/nips/LiB19}.
Our hypothesis is that, if there was no pressure for re-learning, then agents
would fall for the earliest successful strategy and do not consider
alternatives -- stressing the importance of the learnability pressure.

\subsection{Memory constraints}
Human language learning is governed by cognitive constraints such as a limited
memory capacity. These, in turn, affect processes of language evolution and
promote greater convergence to a common language within a community: once
groups become too big, it becomes hard to maintain unique communication
protocols with different partners (i.e.,
idiolects)~\parencite{wray2007consequences}.

Such constraints have been shown to underlie patterns of cross-linguistic
diversity, whereby larger populations develop more structured and less variable
languages \parencite{raviv2019larger}. Yet, neural networks have virtually no
memory constraints because they are commonly heavily over-parametrized. Due to
this over-parametrization, neural networks have no problem to keep a large
number of different partner-specific variants in their memory, and have little
need to converge on a single shared language. However, simply reducing the
number of model parameters to the theoretical minimum is not feasible either,
as explored in emergent communication by
\textcite{DBLP:conf/atal/Resnick0FDC20}. This is because over-parametrization
is, in fact, a critical ingredient for the success of deep neural
networks~\parencite{nakkiran2021deep,aroraFineGrainedAnalysisOptimization2019,zhongRecoveryGuaranteesOnehiddenlayer2017,DBLP:journals/mcss/Cybenko89}.
But given the importance of such memory constraints for human language learning
and evolution, it may be worth considering how such pressures can nonetheless
be mimicked or introduced as inductive biases when employing deep neural
networks as models for language development research.

While large language models have even higher model capacity with billions of learnable parameters, there is an interesing conceptual connection with working memory: As the model parameters are not updated at inference time (when the model is prompted with a specific input), the model can only base its generation on what is available in the prompt, which is limited by the LLMs' context window of how many tokens can be processed at a time. Although also these context windows grow larger and larger with the development of new models~\parencite{gpt4}, it allows researchers to explicitly control what information is available to the model at a specific point in time.

\subsection{Production-comprehension symmetry}
In addition, in naturalistic settings with proficient language users, every person capable of
producing a language is also capable of understanding it
\parencite{hockett1960origin} -- a property that was typically absent from
emergent communication simulations \parencite{galke2022emergent}. Indeed,
introducing an inherent connection between production and comprehension in
neural networks has led to an increase in the desirable properties of emergent
languages~\parencite{michelRevisitingPopulationsMultiagent2023}. Interestingly, comprehension and production are intrinsically linked in autoregressive large language models as the same model parameters are used for processing and for generation~\parencite{radford2019language}. Such results
again underscore the importance of keeping seemingly basic psycholinguistic
features in mind when using large language models and neural networks as models
for human language learning and use. 

\subsection{Modeling other agents' internal states}
Furthermore, another intriguing direction is to explicitly model other agents' internal states.
For instance, \textcite{ohmer2020reinforcement} integrates pragmatic
reasoning into the agents, leading to accelerated learning -- an effect that is
even stronger with Zipfian input distributions compared to uniform input distributions.
Explicitly modeling other agents internal states and social learning has been shown to be successful in other reinforcement
learning scenarios, where agents can cooperate or compete about resources~\parencite{ndousse2021emergent,filos2021psiphi}.
Interestingly, these ideas of explicitly modeling the internal state of the interlocutor 
are already present in the final training stage
of large language models, when optimizing for human
preferences via RLHF~\parencite{schulman2017proximal,rlhf}: the common procedure is to
learn a specific reward model that estimates human preferences on new data, which is then
be employed for steering the generations of the language model in a particular direction -- here the reward model is specifically designed to estimate to what extent humans would prefer one generation over the other, which is closely resembles the idea of modeling other agents' (or humans') internal states.

\section{Discussion}
Several important mismatches between humans and neural agents with respect to
language emergence can be explained by the absence of key cognitive and
communicative pressures, such as memory constraints and
production-comprehension symmetry, which drive language evolution. Here, we
demonstrated how including these factors in neural agents can resolve said
mismatches, and lead to more accurate simulations that mimic the settings and
pressures operating during human language learning and use -- and
consequentially resulting in emergent neural communication protocols that are
more linguistically plausible. Notably, additional psycho- and sociolinguistic
factors may affect language evolution and learning, and might also play a role
in explaining further discrepancies in behavioral patterns across learning
systems. 
 
In the current paper we presented a number of initial mismatches between
humans and agents engaging in communication games -- and demonstrated how they could be
resolved through inductive biases. So far, there is no unified approach that consolidates all of the resolutions mentioned above. We deem
this a promising direction of future work -- \eg, merging the techniques of
population heterogeneity, laziness and impatience, and sender-receiver ties,
which have so far only been evaluated independently.
 
As exemplified by recent work, it is promising to keep up and nourish the
knowledge exchange between researchers working on human languages and those
working on computational simulations of language, \eg, via theory diffusion
from language studies into machine learning and vice versa. A famous example
is cultural
evolution~\parencite{tomaselloOriginsHumanCommunication} and the iterated learning
paradigm \parencite{kirby2014iterated,kirby2008cumulative}, which sparked the
idea of iteratively training neural networks while resetting some of the
networks’
parameters~\parencite{nikishinPrimacyBiasDeep2022,zhouFortuitousForgettingConnectionist2022,DBLP:conf/nips/LiB19,frankleLotteryTicketHypothesis2018a}.
This idea has, for instance, advanced our understanding of neural networks
(their reliance on sparse sub-networks) and led to favorable learning dynamics
that cause better and more systematic generalization beyond the training distribution. 
Similarly, the discrete and compositional structure of natural languages
inspired researchers to incorporate discrete representations into neural
network architectures in order to advance the models' generalization
performance and continual learning
capabilities~\parencite{liuDiscreteValuedNeuralCommunication2021,traubleDiscreteKeyValueBottleneck2023a}.

In conclusion, The emergent communication literature provided
the opportunity to assist in developing linguistic theories in the spirit of
\textcite{elmanLearningDevelopmentNeural1993}, while, conversely, reflecting on
how phenomena and biases known from humans may ultimately enhance neural
networks, as in lifelong and open-world learning, which is still a major open
problem in machine learning.
For making use of large language models in language 
development research, we consider it a promising direction for future work to
take inspiration from the emergent communication literature, and see which
inductive biases (such as the ones sketched here) have helped to recover
patterns from human language learning. Concretely, this would entail
ingesting a training objective for communicative success earlier in
language model training, and integrating a pressure to keep utterances as short as
possible. Integrating these biases into
large language models may very well lead to more cognitively plausible models for
gaining new insights on how children acquire their first language.

\printbibliography

\section{Data, Code and Materials Availability Statement}
This review paper does not introduce any new data, code, or materials.

\section{Authorship and Contributorship Statement}
LG conceptualized the idea, reviewed the literature and wrote the paper. LR conceptualized the idea and helped write the paper. 
 
\section{Acknowledgements}
We thank Mitja Nikolaus and Mathieu Rita for insightful
comments and discussions. We thank Eva Portelance and Michael C Frank for
their valuable comments on an initial version of the manuscript.

\section{License}
Language Development Research (ISSN 2771-7976) is published by TalkBank and the
Carnegie Mellon University Library Publishing Service. Copyright © 2024 The
Author(s). This work is distributed under the terms of the Creative Commons
Attribution Noncommercial 4.0 International license
(\url{https://creativecommons.org/licenses/by-nc/4.0/}), which permits any use,
reproduction and distribution of the work for noncommercial purposes without
further permission provided the original work is attributed as specified under
the terms available via the above link to the Creative Commons website.

\end{document}